\documentclass{article}

\PassOptionsToPackage{numbers, compress}{natbib}

\usepackage[preprint]{neurips_2024}

\usepackage[utf8]{inputenc} 
\usepackage[T1]{fontenc}    
\usepackage{hyperref}       
\usepackage{url}            
\usepackage{booktabs}       
\usepackage{amsfonts}       
\usepackage{nicefrac}       
\usepackage{microtype}      
\usepackage{xcolor}         
\usepackage{colortbl}
\usepackage{multirow}
\usepackage{pifont}
\newcommand{\cmark}{\ding{51}}%
\usepackage{graphicx}
\usepackage{enumitem}
\definecolor{Gray}{gray}{0.9}

\title{Beyond Raw Videos: Understanding Edited Videos with Large Multimodal Model}

\author{%
  Lu Xu$^{\dagger}$, \quad Sijie Zhu$^{\dagger*}$, \quad Chunyuan Li, \quad Chia-Wen Kuo, \quad Fan Chen \AND
  Xinyao Wang, \quad Guang Chen, \quad Dawei Du, \quad Ye Yuan, \quad Longyin Wen \\\\ ByteDance Inc., San Jose, CA, USA
}

\begin{document}

\maketitle

\let\thefootnote\relax\footnotetext{$\dagger$ Equal contribution}
\let\thefootnote\relax\footnotetext{$*$ Corresponding author, sijiezhu@bytedance.com}

\begin{abstract}

The emerging video LMMs (Large Multimodal Models) have achieved significant improvements on generic video understanding in the form of VQA (Visual Question Answering), which mainly focuses on raw videos captured with cameras. However, a large portion of videos in real-world applications are edited videos, \textit{e.g.}, users usually cut and add effects/modifications to the raw video before publishing it on social media platforms. The edited videos usually have high view counts but they are not covered in existing benchmarks of video LMMs, \textit{i.e.}, ActivityNet-QA, or VideoChatGPT benchmark. In this paper, we leverage the edited videos on a popular short video platform, \textit{i.e.}, TikTok, and build a video VQA benchmark (named EditVid-QA) covering four typical editing categories, i.e., effect, funny, meme, and game. Funny and meme videos benchmark nuanced understanding and high-level reasoning, while effect and game evaluate the understanding capability of artificial design. Most of the open-source video LMMs perform poorly on the EditVid-QA benchmark, indicating a huge domain gap between edited short videos on social media and regular raw videos. To improve the generalization ability of LMMs, we collect a training set for the proposed benchmark based on both Panda-70M/WebVid raw videos and small-scale TikTok/CapCut edited videos, which boosts the performance on the proposed EditVid-QA benchmark, indicating the effectiveness of high-quality training data. We also identified a serious issue in the existing evaluation protocol using the GPT-3.5 judge, namely a "sorry" attack, where a sorry-style naive answer can achieve an extremely high rating from the GPT judge, e.g., over 4.3 for correctness score on VideoChatGPT evaluation protocol. To avoid the "sorry" attacks, we evaluate results with GPT-4 judge and keyword filtering. The dataset is released at \url{https://github.com/XenonLamb/EditVid-QA}.

\end{abstract}

\section{Introduction}
\label{sec:intro}
Video has become the major form of media for daily information sharing, while video understanding \cite{survey} remains challenging due to the highly diverse and complex video content in real-world applications. The rapid expansion of social media platforms has not only accelerated the growth of online videos but also introduced new challenges, i.e., understanding artifactual patterns (e.g., effects) or high-level concepts (e.g., funny) requires strong background knowledge and reasoning ability. 

Recent advancement of video LMMs (Large Multimodal Models) \cite{liu2024visual,lin2023video,maaz2023video,li2023llama} shows exciting zero-shot performance on video VQA (Visual Question Answering) with high potential for understanding edited videos. However, the video LMMs are primarily benchmarked \cite{maaz2023video} on regular videos and none of the existing datasets contains evaluation or training data for understanding typical edited videos on social media platforms in the form of VQA likely due to three challenges. 1) It is difficult to generate accurate VQA ground truth for evaluation of edited videos, e.g., even powerful industrial LMMs like GPT-4V \cite{gpt4v} perform poorly on some video categories (Table \ref{tab:sorry}). Some concepts like "funny" and "meme" are challenging even for human beings. 2) Given the model prediction and ground-truth answer, it is hard to evaluate the performance with automatic metrics. The predominant evaluation metrics \cite{maaz2023video} in video LMM literature are the accuracy and score in the range of [0, 5] generated by the GPT-3.5 judge. However, we find that a naive sorry-style answer, denoted as "sorry attack", achieves extremely high scores on most GPT-3.5-based metrics (Table \ref{tab:sorry}), which often occurs in the prediction of GPT-4V. GPT-3.5 judge might be biased toward certain patterns of answers which could be used as a shortcut for other models. This issue can be mostly addressed by using GPT-4 as the judge. 3) It is expensive and challenging to generate large-scale instruction-following data for edited videos, as common annotators are not trained with the required background expertise. Industrial LMMs like GPT-4V also have a relatively low accuracy on some video categories. \textit{The three challenges for building benchmarks for edited videos remain untouched in previous LMM literature.}


In this paper, we propose a new video VQA benchmark, EditVid-QA, as a complement to existing video LMM benchmarks toward understanding typical edited videos on social media platforms. We address the three challenges by collecting new evaluation data, building rectified evaluation metrics, and generating new training data. 1) We focus on four popular categories of edited videos in the evaluation set, i.e., effect, funny, meme, and game. Source videos for "effect" are manually rendered with an off-the-shelf effect tool and raw videos from ImageNet-Vid \cite{deng2009imagenet,shang2017video}. The other categories are public videos collected from a popular social media platform, i.e., TikTok. The evaluation set is relatively small but fully annotated by editing experts with GPT-4V assistance so that the ground-truth question-answer (QA) pairs are highly accurate. 2) To avoid the sorry attack or potential bias of GPT-3.5 judge, we adopt GPT-4 as judge and apply keyword filtering to remove all the answers with "sorry" or "apologize". We benchmark state-of-the-art methods on both the proposed EditVid-QA and VideoChatGPT evaluation set with the rectified metrics. We observe inconsistency in relative performance between different methods, implying the necessity of improving the evaluation metrics. 3) To further boost the performance on the EditVid-QA dataset, we adopt GPT-4V to generate a high-quality training set with source videos from existing datasets, i.e. WebVid \cite{Bain21} and Panda-70M \cite{chen2024panda}. We also manually annotate a small-scale training set with GPT-4V assistance using similar TikTok videos as the evaluation set with no overlap. Experiments show a significant performance boost on EditVid-QA benchmark when extra training data is used in addition to existing training data, i.e., VideoChatGPT \cite{maaz2023video}. We summarize our contributions as follows:
\setlist{nolistsep}
\begin{itemize}[noitemsep,leftmargin=*]
\item We propose a new benchmark for understanding edited videos, i.e., EditVid-QA, as a complement to existing video LMM benchmarks. It contains high-quality evaluation and training data to facilitate the research about edited videos from social media.\\
\item We observe a serious issue with the existing GPT-3.5-based evaluation protocol and propose an alternative solution. We benchmark state-of-the-art methods on both the proposed EditVid-QA and VideoChatGPT datasets with the rectified evaluation metrics.\\ 
\end{itemize}

\begin{table}[!tbp]
    \centering
    \resizebox{\linewidth}{!}{
    \begin{tabular}{l c c c c c c c c c c c}
    \toprule
    \multirow{2}{*}{Model} & \multirow{2}{*}{GPT Judge} & \multicolumn{4}{c}{EditVid-QA} & \multicolumn{5}{c}{VideoChatGPT} \\
    \cmidrule(r){3-6} \cmidrule(r){7-11}
    & & Effect & Funny & Meme & Game & CI & DO & CU & TU & CO \\
    \midrule
    LLaMA-VID \cite{li2023llama} & GPT-3.5 & 15.9 / 2.0 & 31.9 / 2.4 & 21.6 / 2.2 & 36.0 / 2.4 & 2.8 & 2.9 & 3.1 & 2.5 & 2.6 \\
    GPT-4V-Azure \cite{gpt4v} &  GPT-3.5   & 55.4 / 3.2 & 66.9 / 3.2 & 89.1 / 4.0 & 82.9 / 3.6 & 3.6 & 3.3 & 3.4 & 3.4 & 3.3\\
    Sorry Attack & GPT-3.5 & 38.1 / 1.6 & 56.6 / 2.4 & 57.3 / 2.4 & 64.0 / 2.7 & 4.4 & 3.2 & 3.7 & 4.7 & 0.1 \\
    \midrule
    LLaMA-VID \cite{li2023llama} & GPT-4& 6.6 / 0.59 & 29.0 / 1.6 & 15.4 / 1.1 & 11.5 / 0.9 & 2.2 & 2.2 & 2.6 & 1.9 & 2.3 \\
    
    GPT-4V-Azure \cite{gpt4v} & GPT-4 & 44.6 / 2.3 & 49.3 / 2.5 & 80.6 / 4.1 & 61.8 / 3.0 & 2.5 & 2.6 & 2.8 & 1.6 &  2.6  \\
    Sorry Attack & GPT-4 & 0.0 / 0.0 & 0.0 / 0.0 & 0.0 / 0.0 & 0.0 / 0.0 & 0.02 & 0.01 & 0.02 & 0.01 & 2.12\\
    \bottomrule
    \end{tabular}}
    \caption{The performance on the proposed EditVid-QA and VideoChatGPT \cite{maaz2023video} benchmarks w/ different GPT judges. An example of "Sorry Attack" answer is :"Sorry, I can't help with identifying or making assumptions about content in videos". The abbreviations CI, DO, CU, TU, CO denote correctness of information, detail orientation, context understanding, temporal understanding, and consistency. The performance is reported in the form of $x/y$ indicating accuracy and score from GPT judge.}
    \label{tab:sorry}
\end{table}



\section{Related Work}
\label{sec:related}
\textbf{Video Understanding.} Early video understanding literature mainly focuses on action recognition \cite{soomro2012ucf101,kuehne2011hmdb}, which classifies the input video frames with hand-crafted features \cite{lindeberg2012scale} or neural networks \cite{he2016deep,hara2017learning,feichtenhofer2019slowfast}. Recent works apply large-scale pre-training to learning generic video representation \cite{feichtenhofer2022masked, pan2021videomoco} so that they can be fine-tuned to tackle diverse downstream tasks. \textit{However, every task requires specific training data for fine-tuning, which is hard to generalize to real-world scenarios}.
\noindent\textbf{Large Multimodal Models.} Recent advancements in large multimodal models \cite{gpt4v,liu2024visual} show high potential for tackling diverse understanding tasks with one model in a zero-shot inference paradigm. One of the pre-dominant open-source LMMs, LLaVA \cite{liu2024visual}, applies visual instruction tuning on pre-trained vision encoders and LLMs with a high-quality dataset. VideoChatGPT \cite{maaz2023video} proposes a 100K video instruction tuning data and a spatial-temporal pooling architecture for video LMMs. The training data is created using videos from ActivityNet \cite{caba2015activitynet} with manually annotated captions, and GPT-3.5 is adopted for the final QA pairs generation. Video-LLaVA \cite{lin2023video} extends LLaVA for video understanding by adopting the LanguageBind \cite{zhu2023languagebind} as a pre-alignment for image and video encoders. VideoChat \cite{li2023videochat} proposes 11K video instruction data for detailed description and conversion based on GPT-4. LLaMA-vid \cite{li2023llama} proposes to extract only two tokens from each frame with context information from the text query. The recent MVBench \cite{li2023mvbench} proposes a new evaluation benchmark for video LMMs in the form of multiple choice QA and a strong video LMM named VideoChat2. 
\textit{Predominant evaluation protocols for video LMMs are still based on GPT judge and video VQA benchmarks, e.g., VideoChatGPT \cite{maaz2023video}, ActivityNet-QA \cite{yu2019activitynet}, MSVD-QA \cite{xu2017video}, MSRVTT-QA \cite{xu2017video}, TGIF-QA \cite{jang-IJCV-2019}, which do not cover the emerging edited videos on social media.}
\noindent\textbf{Edited Video Dataset} Little effort has been made to understand edited videos and only several datasets are available for academic research in this field. Jafarian et al. \cite{jafarian2021learning} propose TikTok dataset with 300 dancing videos and human masks. AutoTransition \cite{shen2022autotransition} collects 35k transition videos from public video templates on social media platforms. Recently, TikTokActions \cite{qian2024tiktokactions} collected over 28K TikTok videos for human action recognition. Edit3K \cite{gu2024edit3k} has collected a set of rendered videos for understanding 6 types of video editing components. \textit{However, none of the existing datasets contain VQA annotation for popular edited video categories on social media.}\\

\section{EditVid-QA Benchmark}
\label{sec:dataset}
\begin{figure}[tbp]
    \centering
    \includegraphics[width=\linewidth]{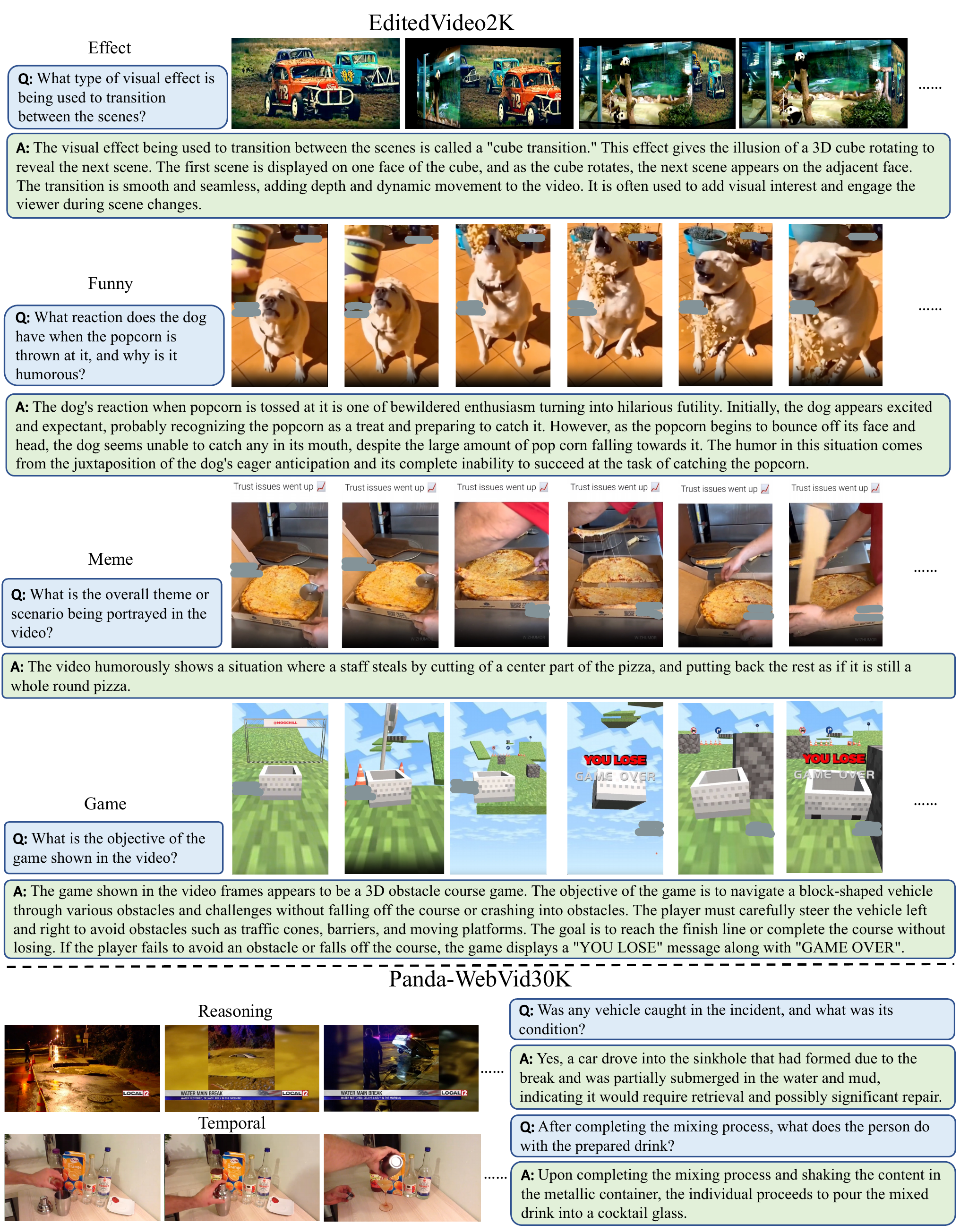}
    \caption{Example video frames and QA pairs of the proposed EditVid-QA dataset. Watermarks are removed for anonymity. Best viewed on the screen with zoom-in.}
    \label{fig:dataset}
\end{figure}

The videos in EditVid-QA are collected from two sources: 1) about 2K videos from TikTok public videos and CapCut rendered videos, named EditedVideo2K. 2) 30K videos from Panda-70M \cite{chen2024panda} and WebVid \cite{Bain21}, named Panda-WebVid30K. The data distribution is shown in Table \ref{tab:dataset}. We also include qualitative examples for each data source in Fig. \ref{fig:dataset}.

\subsection{EditedVideo2K}
\label{sec:tiktok2k}
We leverage short video posting and editing apps to download and render public videos from the internet for 4 popular video editing categories, i.e., effect, funny, meme, and game.


\textbf{Effect.} Visual effects are very popular among edited videos, but multiple effects or filters could be applied to the same videos, which makes it hard for annotators to create ground-truth QA pairs. To avoid data cleaning for online videos, we adopt an off-the-shelf editing tool, i.e., CapCut \cite{capcut}, to render videos with various effects. We use ImageNet-vid \cite{deng2009imagenet} videos as raw videos and each rendered video only contains one effect, including video effect, animation, transition, and filter. The relatively simple background context and the single effect setting make it easier for annotators to create ground-truth QA pairs. \textit{The major challenge for effect videos it that the model needs to distinguish the visual effect and the background content in the input frames, which is not considered in previous works.} \\
\textbf{Funny.} Funny short videos are very popular on social media platforms. We select a group of TikTok \cite{tiktok} video creators with over 1M followers and collect around 1K videos with "funny" in the hashtag. Most of the funny points lie in the vision content, e.g., a funny-looking pet, an interesting movement, a weird gesture, etc. \textit{It might be easy to generate descriptions or answer factual questions for these videos, but funny points usually require background knowledge and reasoning to understand, which could be challenging for existing LMMs.}
\begin{table}[tbp]
 \caption{The video and question-answer (QA) statistics of the proposed EditVid-QA benchmark.}
     \centering
     \begin{tabular}{l l c c c c c c}
     \toprule
     & & \multicolumn{4}{c}{\textit{EditedVideo2K}} & \multicolumn{2}{c}{ \textit{Panda-WebVid30K}}\\
     \cmidrule(r){3-6} \cmidrule{7-8} 
     & & Effect & Funny & Meme & Game & Reasoning & Temporal\\
     \midrule
    \multirow{2}{*}{Evaluation} & \#Video &  99 & 101 & 100 & 62 & - & - \\
    & \#QA & 121 & 138 & 104 & 76 & - & - \\
    \midrule
    \multirow{2}{*}{Training} & \#Video & 870 & 420 & 520 & 437 & 29,842 & 9,842 \\
    & \#QA & 3,320 & 1,680 & 1,924 & 1,723 & 145,500 & 49,210 \\
    \bottomrule
     \end{tabular}
     \label{tab:dataset}
     \vspace{-0.2cm}
\end{table}

\noindent\textbf{Meme.} Meme videos are also funny but in a different way. It usually contains text overlay or emoji as the major storyline or theme of the video, which often reflects the thoughts/feelings of the audience or involves ironic jokes. The visual content itself may not be funny, but the overall ironic point can easily go viral on social media platforms. We ask annotators to manually select 1K TikTok \cite{tiktok} meme videos from a candidate pool. \textit{To understand meme videos, the model must have outstanding OCR performance to catch the major theme and connect the text theme to the visual content based on background knowledge.} \\
\textbf{Game.} Game videos are usually captured with a phone camera with game tools following manually designed game logic, e.g., control the motorcycle with nose or phone pose so that it drives to destination. These videos look dramatically different from regular videos and the game logic requires strong background knowledge to understand from visual input. Also, one often needs to watch till the end of the video to get the overall logic of the game, e.g., win/lose or the score. We collect 800 TikTok \cite{tiktok} videos with popular game tools and filter out the invalid videos with the wrong category or low quality. \textit{The major challenge is the limited visual clue, i.e., there is no written game rule in the video and the models need to consider a wide range of frames to understand the game logic.}\\

\textbf{Annotation.} Given the extracted frames of each video, we first adopt GPT-4V to generate 5 questions related to the corresponding category with an in-context example, e.g., a typical question for "effect" is: "What visual effect is applied in this video?" Detailed example prompts are included in the supplementary material. Then the answer is generated with GPT-4V using the question and video frames as input. We find that GPT-4V answers have extremely high accuracy on "meme" videos, but the accuracy is relatively low on the other three categories. 

As for the evaluation set, we randomly select around 100 videos from each category and ask annotators to rewrite the incorrect answers. If all the QAs are wrong, the annotators will create at least one QA pair using a list of template questions (examples included in the supplementary material). The rest videos are adopted for training and the annotators select all the correct QA pairs from GPT-4V without rewriting, resulting in a small-scale training set. The size of this dataset is relatively small, but it provides valuable insight into the benefit of using edited videos as training.
 
\subsection{Panda-WebVid30K}
\label{sec:Panda-WebVid}
Most of the previous works are trained on the 100K data from VideoChatGPT \cite{maaz2023video} which is created with GPT-3.5 with more risk of hallucination than GPT-4V. We take advantage of GPT-4V to create a set of high-quality instruction-tuning data, i.e., Panda-WebVid30K, as a good complement to existing datasets. Instead of focusing on factual questions, our training data is created with two types of prompts, i.e., reasoning and temporal. The reasoning QAs pairs help the model generalize on edited videos which involves understanding high-level concepts. Temporal QAs pairs enhance the ability of theme/logic understanding from a wide range of frames.

The source videos consist of 20K videos randomly sampled from the WebVid dataset and 9.8K videos from Panda70M dataset which are selected based on a maximum duration of 45 seconds. We extract 10 frames uniformly from each video as the visual input of GPT-4V and generate 5 QA pairs. Both WebVid and Panda70M source videos are used for generating reasoning QA pairs, and only Panda70M is used for temporal QA pairs because WebVid videos are mostly single-action shots.


\begin{figure}[!htbp]
    \centering
    \includegraphics[width=\linewidth]{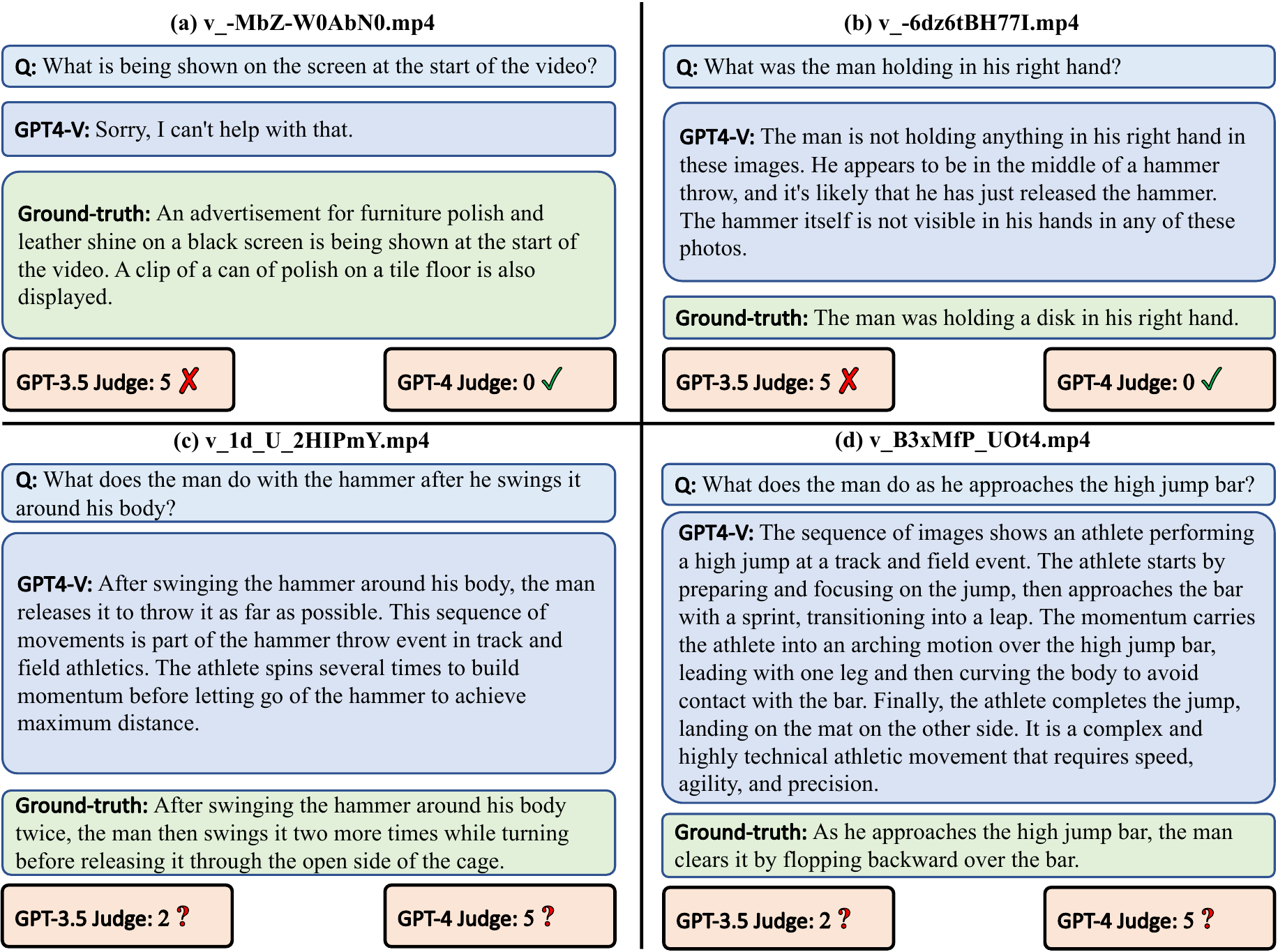}
    \caption{Comparison between GPT-3.5 and GPT-4 judge on VideoChatGPT dataset \cite{maaz2023video}. The two GPT judges could be inconsistent for some cases. }
    \label{fig:evaluation}
\end{figure}

\subsection{Evaluation Metrics}
\label{sec:evaluation}
The predominant evaluation protocol is proposed in VideoChatGPT \cite{maaz2023video}, where model predictions and ground-truth answers are sent to GPT-3.5 judge to generate a score in [0,5]. However, we find that GPT-3.5 judge has a bias toward certain patterns of answers, e.g., a sorry-style answer (see examples in Fig. \ref{fig:evaluation} (a)). Table \ref{tab:sorry} shows that a naive sorry answer can significantly outperform existing LMMs on the correctness score of VideoChatGPT \cite{maaz2023video} benchmark, while GPT-4 judge can solve the sorry-style bias for most categories. Fig. \ref{fig:evaluation} (b) shows that GPT-3.5 judge could also be biased toward negative answers, e.g., claiming there is no object. Sometimes the GPT-3.5 judge would put punishment on providing more details than ground-truth answer (Fig. \ref{fig:evaluation} (c), (d)), which is correct to some extent. However, GPT-4 judge tends to give a high score for such detailed answers based on background knowledge. Overall, we observe that GPT-4 has better judgment than GPT-3.5 on scoring the correctness of predictions with given ground-truth answers.

To establish more convincing evaluation metrics, we follow the previous evaluation prompts but replace the GPT-3.5 with GPT-4 as the judge and add keyword filtering for sorry-style keywords. We find that the results from GPT-3.5 and GPT-4 judge could be very different for the same method, and the ranking of different methods could also be inconsistent (Table \ref{tab:academic} in Sec. \ref{sec:benchmarking}), indicating the necessity of changing GPT judge. 





\section{Experiment}

\subsection{Implementation Details}

All the models are implemented with Pytorch \cite{paszke2019pytorch} and trained with AdamW \cite{loshchilov2017decoupled} optimizer. Mistral-Ins-0.2 \cite{jiang2023mistral} is adopted as LLM and CLIP \cite{clip} is used as the vision backbone. For the pre-training stage of LLaMA-VID models, we only update the MLP projector with the learning rate of 1e-3. In the instruction tuning stage, we update both the projector and the LLM with the learning rate of 1e-6. We train the model for one full epoch using Deepspeed \cite{rasley2020deepspeed} ZeRO-2. For all training videos, we extract the frames at 1 fps (frame per second) and each frame is resized to the resolution of 224$\times$224. The whole training process takes approximately 30 hours on 32 A100 GPUs. We use the Azure API for all the GPT calls, including GTP-3.5 (turbo), GPT-4, and GPT-4V. We use 10 input frames when calling GPT-4V because of the maximum limit of Azure \cite{azure} service.

\subsection{Benchmarking State-of-the-art Methods}
\label{sec:benchmarking}

\begin{table}[!tbp]
  \caption{Performance comparison between the proposed models and state-of-the-art video LMM methods on proposed EditVid-QA evaluation set. The performance is reported in the form of $x/y$ indicating accuracy and score from GPT-4 judge with keyword filtering, i.e., removing invalid answers with "sorry" or "apologize". The same GPT-4 version (0613) is used for all methods in this table. $\dagger$ denotes the reproduced version using Mistral.}
  \label{tab:edit-vqa}
  \centering
  \resizebox{\linewidth}{!}{
  \begin{tabular}{l l c c c c |c}
    \toprule
    Method  & LLM  & Effect & Funny & Meme & Game & Avg.\\
    \midrule
    \rowcolor{Gray}
    GPT-4V(Azure) \cite{gpt4v} & NA & 44.6 / 2.3 & 49.3 / 2.5 & 80.6 / 4.1 & 61.8 / 3.0 & 59.1 / 3.0 \\
    \midrule
    LLaVA-NeXT-Video-7B \cite{llava1.6} & Vicuna-1.5-7B & 23.9 / 1.4 & 29.7 / 1.8 & 26.0 / 1.9 & 28.9 / 1.7 & 27.1 / 1.7\\
    \midrule
    Video-LLaVA \cite{lin2023video} & Vicuna-1.5-7B & 15.7 / 0.9 & 18.8 / 1.3 & 11.5 / 0.9 & 18.4 / 1.2 & 16.1 / 1.1 \\
    Video-ChatGPT \cite{maaz2023video} & Vicuna-7B & 13.2 / 0.8 & 20.3 / 1.3 & 13.5 / 1.0 & 14.5 / 1.2 & 15.4 / 1.1 \\
    VideoChat \cite{li2023videochat} &  Vicuna-7B & 19.8 / 1.0 & 18.8 / 1.2 & 16.3 / 1.2 & 18.4 / 1.3 & 18.3 / 1.2 \\
    VideoChat2 \cite{li2023mvbench} & Vicuna-7B & 24.8 / 1.1 & 21.7 / 1.4 & 11.5 / 1.1 & 18.4 / 1.1 & 19.1 / 1.2 \\
    LLaMA-VID \cite{li2023llama} & Vicuna-1.5-7B & \,\,6.6 / 0.6 & 29.0 / 1.6 & 15.4 / 1.1 & 11.5 / 0.9 & 15.6 / 1.1 \\
    $^{\dagger}$LLaMA-VID \cite{li2023llama}  & Mistral-Ins-0.2-7B & \,\,8.3 / 0.7 & 18.8 / 1.3 & 14.4 / 1.1 & 22.4 / 1.4 & 16.0 / 1.1 \\
    \textbf{$^{\dagger}$LLaMA-VID + Our Data} & Mistral-Ins-0.2-7B  & 13.2 / 0.9 & 34.8 / 1.9 & 18.3 / 1.2 & 46.1 / 2.4 & 28.1 / 1.6 \\  
    \bottomrule
  \end{tabular}}
\end{table}

We benchmark open-source state-of-the-art video LMMs on the EditVid-QA, ActivityNet-QA, and VideoChatGPT benchmarks using the official GitHub repositories, including Video-LLaVA \cite{lin2023video}, Video-ChatGPT \cite{maaz2023video}, VideoChat \cite{li2023videochat}, LLaMA-VID \cite{li2023llama}. We select the 7B configuration for all models for fair comparison. The performance of GPT-4V is also included for reference. To demonstrate the effectiveness of the proposed training data, we adopt LLaVA-VID as our baseline method because of its simple training paradigm, and train it with the proposed data recipe, denoted as "$\dagger$LLaMA-VID + Our Data". Due to potential legal issues of LLaMA2, we replaced the Vicuna 1.5 \cite{zheng2023judging} in LLaMA-VID with Mistral, and the reproduced version is marked with $\dagger$. LLaVA-NeXT \cite{llava1.6} is released very recently with a much better performance than published works due to more advanced training recipe and architecture, but the training code is not available. Other concurrent models are included in supplementary material. 

\begin{table}[!tbp]
  \caption{Performance comparison between the proposed models and state-of-the-art video LMM methods on ActivityNet-QA (AN-QA) datasets and the VideoChatGPT benchmark \cite{maaz2023video}. The abbreviations CI, DO, CU, TU, CO denote the correctness of information, detail orientation, context understanding, temporal understanding, and consistency. The performance on ActivityNet-QA datasets is reported in the form of $x/y$ indicating accuracy and score from GPT-4/3.5 judge with keywords filtering, i.e., removing invalid answers with "sorry" or "apologize". The same GPT-4 (0613) and GPT-3.5-turbo (0301) versions are used for all methods in this table. $\dagger$ denotes the reproduced version using Mistral.}
  \label{tab:academic}
  \centering
  \resizebox{\linewidth}{!}{
  \begin{tabular}{l | l l  c c c c c c }
    \toprule
   \multirow{2}{*}{Judge} & \multirow{2}{*}{Method} & \multirow{2}{*}{LLM} & \multirow{2}{*}{AN-QA} & \multicolumn{5}{c}{VideoChatGPT}\\ 
    \cmidrule(r){5-9} 
    & & & & \footnotesize{CI} & \footnotesize{DO} & \footnotesize{CU} & \footnotesize{TU} & \footnotesize{CO} \\
    \midrule
    \multirow{7}{*}{GPT-4} & GPT-4V(Azure) \cite{gpt4v} &  NA & 43.1 / 2.1 & 2.5 & 2.6 & 2.8 & 1.6 &  2.6 \\
    & Video-LLaVA \cite{lin2023video} & Vicuna-1.5-7B & 42.6 / 2.1 & 2.2 & 2.1 & 2.6 & 1.9 & 2.4 \\
    & Video-ChatGPT \cite{maaz2023video} &  Vicuna-7B & 39.7 / 2.0 &  1.6 & 1.6 & 2.0 & 1.4 & 1.7 \\
    & VideoChat \cite{li2023videochat} &  Vicuna-7B & 29.4 / 1.4 & 1.5 & 1.9 & 1.9 & 1.2 & 1.2 \\
    & LLaMA-VID \cite{li2023llama} & Vicuna-1.5-7B & 42.8 / 2.1 & 2.2 & 2.2 & 2.6 & 1.9 & 2.3 \\
    & $^{\dagger}$LLaMA-VID \cite{li2023llama} & Mistral-Ins-0.2-7B & 39.8 / 2.0 &  1.7 & 1.8 & 2.1 & 1.5 & 1.8 \\
    & \textbf{$^{\dagger}$LLaMA-VID + Our Data}  & Mistral-Ins-0.2-7B & 42.2 / 2.1 &1.8 & 2.1 & 2.2 & 1.6 & 1.6 \\
    \bottomrule
    \multirow{7}{*}{GPT-3.5} & GPT-4V(Azure) \cite{gpt4v} &  NA & 55.4 / 2.8 & 3.0 & 2.8 & 3.1 & 2.1 & 2.7 \\
    & Video-LLaVA \cite{lin2023video} & Vicuna-1.5-7B & 49.7 / 3.5 &  2.2 & 2.1 & 2.6 & 1.9 & 2.5 \\
    & Video-ChatGPT \cite{maaz2023video} & Vicuna-7B & 48.3 / 3.4 &  2.4 & 2.6 & 2.7 & 2.2 & 2.3 \\
    & VideoChat \cite{li2023videochat} &  Vicuna-7B & 44.1 / 2.6 & 1.5 & 1.9 & 1.9 & 1.2 & 1.2 \\
    & LLaMA-VID \cite{li2023llama} & Vicuna-1.5-7B & 50.4 / 3.5 &  2.8 & 2.9 & 3.1 & 2.5 & 2.6  \\
    & $^{\dagger}$LLaMA-VID \cite{li2023llama} & Mistral-Ins-0.2-7B & 47.1 / 3.3 & 2.3 & 2.5 & 2.7 & 2.3 & 2.2 \\
    & \textbf{$^{\dagger}$LLaMA-VID + Our Data} & Mistral-Ins-0.2-7B & 53.5 / 3.3 & 2.4 & 2.7 & 2.9 & 2.0 & 2.2  \\
    \bottomrule
  \end{tabular}}
\end{table}

In Table \ref{tab:edit-vqa}, we benchmark the state-of-the-art video LMMs on the proposed EditVid-QA evaluation set with GPT-4 judge. The industrial model, GPT-4V \cite{gpt4v}, achieves the best performance with a large margin over open-source models. However, the accuracy of GPT-4V on three categories is still lower than $65\%$, indicating the proposed EditVid-QA benchmark is extremely challenging. GPT-4V has a high accuracy on "Meme" categories because of its strong multi-lingual OCR ability and background knowledge. In contrast, open-source 7B models do not generalize well on the four categories of edited videos. The accuracy of existing models is lower than $30\%$ for all categories. Compared to the baseline "$\dagger$LLaMA-VID" model, adding our training data brings a decent performance boost over all 4 categories, especially on "Funny" ($+16\%$) and "Game" ($+23.7\%$). The average performance of "$\dagger$LLaMA-VID + Our Data" is much better than other published open-source models. 

    

We also benchmark the performance of LMMs on academic datasets, i.e., ActivityNet-QA \cite{yu2019activitynet} and VideoChatGPT \cite{maaz2023video}, with both GPT-4 and GPT-3.5 judges in Table \ref{tab:academic} with keywords filtering for sorry-style answers. For GPT-4 judge, we do not observe a significant performance gap between GPT-4V and open-source LMMs. LLaMA-VID, "$^{\dagger}$LLaMA-VID" and "$^{\dagger}$LLaMA-VID + Our Data" performs on par with GPT-4V on ActivityNet-QA, but they have relatively low accuracy on VideoChatGPT benchmark. Adding the proposed training data also shows a consistent performance boost across different metrics.

Another key observation is the inconsistency between GPT-4 and GPT-3.5 judges, i.e., the rank of different models could be different when GPT judge is changed. For example, our model achieves $53.5\%$ accuracy on ActivityNet-QA based on GPT-3.5 judge which is better than other published open-source models, but the GPT-4 judge score is similar to other models, e.g., "LLaMA-VID" and "Video-LLaVA". On the other hand, video-LLaVA shows relatively low accuracy when evaluated with GPT-3.5 judge, but the performance of Video-LLaVA is better than other open-source models on most metrics. Therefore, it is necessary to switch to GPT-4 judge for a more reliable evaluation.

\subsection{Ablation Study}

To demonstrate the effectiveness of the proposed training data, we conduct a detailed ablation study on different training data combinations in Table \ref{tab:data-ablation}. The first three rows show that the combination of VideoChatGPT100K and the proposed Panda-WebVid30K performs much better than using only one training dataset. The performance is improved on all four categories. Since EditedVideo2K contains much fewer videos than other datasets, we adopt it for a final fine-tuning stage by default, i.e., first train with other data and then fine-tune with EditedVideo2K. Intuitively the EditedVideo2K should have a similar distribution as the evaluation videos, but performance improvements for the four categories are quite different. The most significant improvement is observed for the "Game" category, possibly due to the similarity between the logic of different games. The performance improvement on "Meme" and "Funny" are moderate, because these two categories rely on background knowledge and reasoning capability, which is hard to be boosted by small-scale fine-tuning. The yes/no accuracy on "Effect" is slightly lower, but the score is slightly improved. The 2K data does not help the model distinguish the artifactual patterns or geometric transformations in visual effect from the raw video content. 
It is worth exploring scalable training data creation strategies in the future. Overall, adding EditedVideo2K for training is still beneficial for the generalization of LMMs on edited videos. 

\begin{table}[!tbp]
  \caption{Performance comparison of the proposed method with different instruction-tuning datasets. EditedVideo2K is used for an additional final-stage fine-tuning because of its data size. All the models are trained with LLaMA-VID architecture and Mistral-7B. The abbreviations CI, DO, CU, TU, CO denote correctness of information, detail orientation, context understanding, temporal understanding, and consistency. The performance on AC-QA and EditVid-QA is reported in the form of $x/y$ indicating accuracy and score from GPT-4 judge. Here VC100K, PW30K, EV2K denote the VideoChatGPT100K, Panda-Web30K, and EditedVideo2K.}
  \label{tab:data-ablation}
  \centering
  \resizebox{\linewidth}{!}{
  \begin{tabular}{c c c c c c c c c c c c c c}
    \toprule
    \multicolumn{3}{c}{Instruction Data}  & \multicolumn{5}{c}{EditVid-QA} & \multicolumn{5}{c}{VideoChatGPT}\\
     \cmidrule(r){1-3} \cmidrule(r){4-8} \cmidrule(r){9-13}
    \footnotesize{VC100K} & \footnotesize{PW30K} & \footnotesize{EV2K} & \footnotesize{Effect} & \footnotesize{Funny} & \footnotesize{Meme} & \footnotesize{Game} & Avg. & \footnotesize{CI} & \footnotesize{DO} & \footnotesize{CU} & \footnotesize{TU} & \footnotesize{CO} \\
    \midrule 
    \cmark & & & 8.3 / 0.7 & 18.8 / 1.3 & 14.4 / 1.1 & 22.4 / 1.4 & 16.0 / 1.1 & 1.7 & 1.8 & 2.1 & 1.5 & 1.8 \\
     & \cmark &  & 8.3 / 0.6 & 22.5 / 1.5 & 11.5 / 0.9  & 13.2 / 1.0 & 13.9 / 1.0 & 1.7 & 1.7 & 2.0 & 1.6 & 1.6 \\
    \cmark & \cmark & & 14.0 / 0.8 & 30.4 / 1.8 & 13.5 / 1.1 & 25.0 / 1.6 & 20.7 / 1.3 & 1.8 & 1.8 & 2.2 & 1.7 & 1.7 \\
  
    \cmark & \cmark & \cmark & 13.2 / 0.9 & 34.8 / 1.9 & 18.3 / 1.2 & 46.1 / 2.4 & 28.1 / 1.6 & 1.8 & 2.1 & 2.2 & 1.6 & 1.6 \\
    \bottomrule
  \end{tabular}}
\end{table}

\subsection{Qualitative Results}
\begin{figure}[!tbp]
    \centering
    \includegraphics[width=\linewidth]{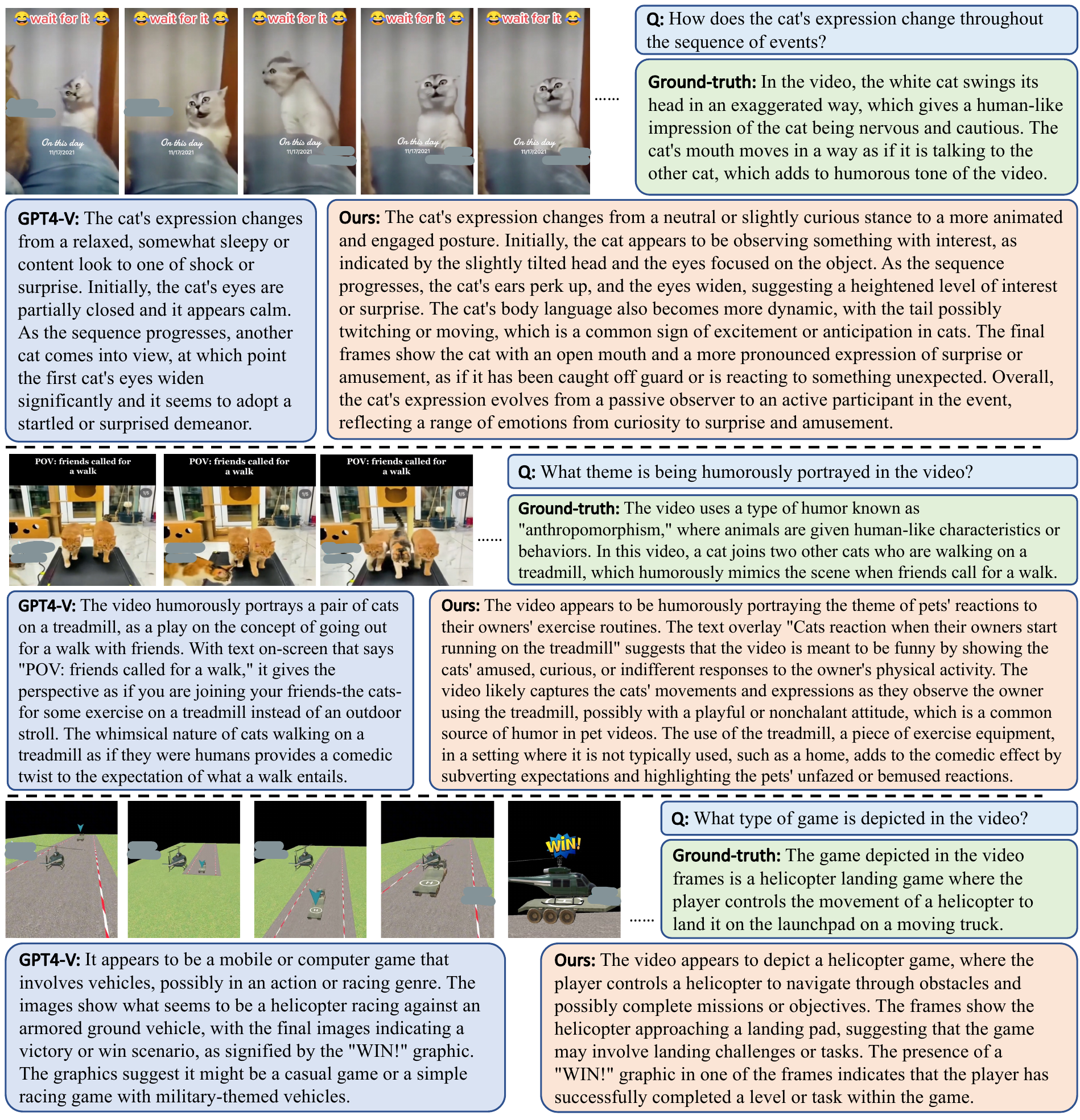}
    \caption{Qualitative results of our model and GPT-4V on the proposed EditVid-QA benchmark. Watermarks are removed for anonymity.}
    \label{fig:qualitative}
\end{figure}

In Fig. \ref{fig:qualitative}, we provide three cases on EditVid-QA benchmark to illustrate the prediction of LLaMA-VID trained with our data, denoted as "Ours". We add the results of GPT-4V for reference. Some cases are challenging even for GPT-4V while our re-trained model provides good answers, e.g., the first and third cases. However, we also observe hallucinations and poor OCR capacity of our model in the second case, i.e., the cat meme video. Both the architecture and training data need to be improved for better generalization performance on edited videos.

\section{Conclusion}
We propose a new video VQA benchmark (EditVid-QA) for LMMs toward better understanding of popular edited videos on social media platforms, as a complement to existing video LMM benchmarks. We also identify a serious issue with the predominant evaluation metric, i.e., GPT-3.5 judge may be biased toward certain patterns. To provide more trustful evaluation metrics, the proposed benchmark adopts GPT-4 as the judge with manually annotated ground-truth QA pairs. We benchmark published open-source LMMs as well as GPT-4V to facilitate future research in this field. In addition, we propose two sets of training data to improve the generalization of LMMs on edited videos, which is shown to be effective with detailed ablation experiments. \\
\textbf{Limitations.} One limitation is that the four categories do not fully cover the edited videos on social media. Other categories like "Montage" or "otoMAD" videos are also popular and will be studied in future work to evaluate the storyline understanding ability of LMMs. Videos directly generated from text-to-video or image-to-video models could also be included in the future. \\
\textbf{Boarder Impacts.} We believe this work will facilitate future research for understanding edited videos with LMMs. It could help social media platforms to better understand online videos, leading to better video recommendation and creation. The authors do not foresee any negative societal impact.

{\small
\bibliographystyle{plain}
\bibliography{main}
}

\end{document}